\documentclass[times, 12pt, a4paper]{article}

\usepackage[utf8]{inputenc}
\usepackage{graphicx}
\usepackage{epstopdf}
\usepackage{subfig}
\usepackage{gensymb}
\usepackage{amsmath}
\usepackage{algorithm}
\usepackage{algorithmic}
\usepackage[usenames, dvipsnames]{color}
\usepackage{authblk}

\newcommand{\locimage}[1]{#1.png}
\providecommand{\keywords}[1]{\textbf{\textit{Keywords---}} #1}

\begin{document}

\title{Learning Representations from Road Network for End-to-End Urban Growth Simulation}

\author[1]{Saptarshi Pal}
\author[1]{Soumya K Ghosh}

\affil[1]{(Department of Computer Science and Engineering, Indian Institute of Technology Kharagpur, India.)}

\maketitle

\begin{abstract}

From our experiences in the past, we have seen that the growth of cities is very much dependent on the transportation networks. In mega cities, transportation networks determine to a significant extent as to where the people will move and houses will be built. Hence, transportation network data is crucial to an urban growth prediction system. Existing works have used \textit{manually derived distance based features} based on the road network to build models on urban growth. But due to the non-generic and laborious nature of the manual feature engineering process, we can shift to \textit{End-to-End} systems which do not rely on manual feature engineering. In this paper, we propose a method to integrate road network data to an existing \textit{Rule based End-to-End} framework without manual feature engineering. Our method employs recurrent neural networks to represent road network in a structured way such that it can be plugged into the previously proposed \textit{End-to-End} framework. The proposed approach enhances the performance in terms of \textit{Figure of Merit, Producer's accuracy, User's accuracy} and \textit{Overall accuracy} of the existing \textit{Rule based End-to-End} framework.  

\end{abstract}


\section{Introduction}

Urbanization is one of the major issues which has been recognized by the United Nations. The United Nations World Urbanization Prospects have pointed out the imminent rise in the urban population which is expected in the coming years\footnote{http://www.un.org/en/development/desa/publications/2014-revision-world-urbanization-prospects.html}. This has led to a growing concern as to how cities can be planned such that uncontrolled urban growth can be controlled.

One of the solutions which have been proposed in the previous works is to build a simulator which can learn patterns from past growth and simulate the future conditions (\cite{moghadam2013spatiotemporal, shafizadeh2015performance, liu2008modelling}). Significant work has been carried out in simulating urban growth in many major cities in the world. The problem though is that the process of simulation is laborious and non-generic due to manual feature engineering from various data sources (\cite{pal2017rule}). The manual feature engineering process mainly consists of generating distance based features, for instance, distance from roads, distance from railways etc.

In order to reduce the dependency on \textit{manualy derived distance based features}, a \textit{Rule based End-to-End} framework have been proposed in \cite{pal2017rule}. The \textit{Rule based End-to-End} framework uses satellite imagery and urban built-up maps to learn patterns of urban growth and simulate future growth. The work (\cite{pal2017rule}) also shows empirically that the proposed \textit{End-to-End} framework provides better performance measure than the existing learning based methods.

Transportation networks are a prominent feature in an urban landscape and people generally find it convenient to build establishments considering the availability of roads or railways. Hence, it can be expected that a road network data is an important constituent in urban growth prediction model. Moreover, as a data source, plenty of crowd sourced as well as legacy data of roads are available online and with government organizations. Therefore, it is important to make use of such data to refine the process of urban growth modeling in an \textit{End-to-End} fashion.

As of now, the \textit{Rule based End-to-End} framework uses only satellite raster image as data source without feature engineering to simulate urban growth. Therefore, to maintain consistency in the framework it is also required that the road network be incorporated in a way which does not require feature engineering. Transportation networks have been used in several other works in deriving distance based features. \textit{Distance from roads or highways} are an important feature which has been considered in the works by \cite{moghadam2013spatiotemporal, shafizadeh2015performance, lin2011predictive, musa2016review}. In these works, only a particular distance based metric (euclidean) has been used to compute distance. It is true that distance from roads determines urban growth but this may not necessarily mean it has to be a particular distance metric (say euclidean distance). The use of a particular distance metric reduces the generic nature of the model by making it rigid. Hence, we intend to use road network in a fashion which does not require computing a fixed manually selected distance metric.

One of the primary challenges in incorporating road network data is due to the fact that roads are represented in vector form and the \textit{Rule based End-to-End} framework is designed to take input raster images only. This is because the framework is based on cellular automata which assumes space as a discrete array of cells and raster representation of data matches well with this assumption. However, road network is represented in terms of polylines consisting of sequences of coordinates with no assumption akin to discrete array of cells in a cellular automata. Therefore, we need to find a way to represent road network so that they can be incorporated into the \textit{Rule based End-to-End} framework without reducing the performance of the simulation.

Our proposed method utilizes concepts of recurrent neural networks (\cite{haykin2009neural}) and autoencoders (\cite{vincent2010stacked}) to represent road network in an unsupervised way. In more common terms, this is referred to as \textit{Representation learning} (\cite{bengio2013representation}), which involves automatic discovery of representations for feature detection, classification or model building purposes. Representation learning have been utilized in several works starting from image recognition (\cite{badrinarayanan2015segnet}), video sequences (\cite{srivastava2015unsupervised}), speech sequences (\cite{graves2013speech, graves2014towards}), text sequences (\cite{sutskever2014sequence}), etc for achieving higher performances. Since road network are sequences of coordinates similar to texts, video, and speech, hence one can attempt to represent them using recurrent neural networks.

In this paper, we propose a novel method of integrating road network into the \textit{Rule based End-to-End} framework. Our proposed solution utilizes the fact that road network in vector form is a sequence of points and hence can be represented using recurrent neural networks. Following the representation, we have successfully integrated the road network representation into the \textit{End-to-End} framework and have achieved superior performance. Our main contributions in the work are given as follows.

\begin{itemize}

\item Utilizing recurrent neural networks to develop a method of representing road network in an \textit{End-to-End} fashion (without feature engineering).

\item Successfully integrating the generated representation into the \textit{End-to-End} framework while achieving superior performance over the existing framework.

\item Additionally, we also show that our proposed method performs better than a naive baseline method for incorporating road network to the \textit{End-to-End} framework.

\end{itemize}

The rest of the sections are organized as follows. In section $2$, we explain the preliminaries of the methods used in the proposed methods. Section $3$ describes our proposed methods of representing road network along with the procedures. Section $4$ presents results and discussions of the experiments conducted in the region of Mumbai, India. Finally, section $5$ provides conclusion and future research directions.

\section{Preliminaries}

In this section, we discuss the \textit{Rule based End-to-End Learning framework}, Recurrent Neural Networks (RNN), Long Short Term memory (LSTM) cells and autoencoders. We also provide a naive baseline method to incorporate road network data into \textit{Rule based End-to-End} framework in order to show the effectiveness of our method.

\subsection{Rule Based End-to-End framework for Urban Growth Prediction} \label{sec:endtoend} 

The Rule based \textit{End-to-End} framework (\cite{pal2017rule}) is a supervised learning framework based on cellular automata model which uses satellite raster data to predict urban growth without \textit{manually selected distance based features}. The idea is to reduce the dependency on fixed distance metrics from the modeling process and provide better results in terms of four metrics namely \textit{Figure of Merit, Producer's accuracy, User's accuracy} and \textit{Overall accuracy} (\cite{pontius2008comparing}). 

The framework utilizes satellite raster and builtup images to develop a cellular automata (CA) model. The built-up images consists of built-up information at time $t$ (represented by $l_p^t$) and has the same resolution as the raster image. The urban growth CA model can be described using a state ($S_t^p$), a neighborhood criterion ($N$), a transition function (eqn. \ref{eqn:ca_model_update}) and an update function (eqn. \ref{eqn:ca_model_transition}). The state of the cellular automata at point $p$ and time $t$ is defined as a triple given as $<l_p^t, \tau_p^t, R_p>$, where 

\begin{itemize}

\item $l_p^t$ is a binary variable representing \textit{Builtup} ($S_B$) or \textit{Non-Builtup} ($S_{NB}$), 

\item $\tau_p^{t}$ is a transition indicator which can take four indicators namely \textit{Non Built-up} to \textit{Non Built-up} ($C_{NB}^{NB}$), \textit{Built-up} to \textit{Built-up} ($C_{B}^{B}$), \textit{Non Built-up} to \textit{Built-up} ($C_{NB}^{B}$) and \textit{Built-up} to \textit{Non Built-up} ($C_{B}^{NB}$) and

\item $R_p$ represents raster variable at point.

\end{itemize}

The generalized form of the two functions (update and transition) in the \textit{End-to-End} framework are given as follows.

\begin{eqnarray}
\tau_p^{t+1} = f_T(l_p^t, N(l_p^t), \phi_{raster}^{len}(R_p, N(R_p)))
\label{eqn:ca_model_update}
\end{eqnarray}

The function $\phi_{raster}^{len}$ is a encoding function to encode neighborhood information of a cell into a fixed length representation of size $len$. This is important because satellite raster data is an unstructured source of data which needs to be converted to a structured form in order to fit into a supervised learning model. The encoding function $\phi_{raster}^{len}$ is developed using a specialized neural architecture known as an autoencoder \cite{pal2017rule}.

\begin{eqnarray}
S_p^{t+1} = 
\begin{cases}
(S_B, \tau_p^{t+1}, R_p) & \tau_p^{t+1}\in \{C_{NB}^B,C_B^B\} \\
(S_{NB}, \tau_p^{t+1}, R_p) & \tau_p^{t+1}\in \{C_{NB}^{NB},C_{B}^{NB}\}
\end{cases} 
\label{eqn:ca_model_transition}
\end{eqnarray}

The important aspect to determine in eqn. (\ref{eqn:ca_model_update}) is a way to build function $f_T$ so that enhanced performance in terms of $FoM, PA, UA$ and $OA$ can be achieved. In order to do so, the framework involves two stages, namely \textit{Data Representation} which include formation of a data and a label matrix, and \textit{Knowledge Representation} which includes training a classifier on the data and label matrices. The trained classifier has been considered as a knowledge representation of the urban growth pattern (\cite{pal2017rule}). The framework is called \textit{Rule based} as experiments have shown that Decision tree and Ensembles of Decision tree have provided better results when compared to other classifiers.

One of the drawbacks of the \textit{End-to-End} framework is that it is based on a cellular automata model. Since initial assumption of a CA model includes a discrete array of cells, hence raster data with same resolution can be easily integrated into the framework. However, many datasets such as road network, railways etc. happen to be in vector form and are not easily integrable to the framework. This limits the utility of the framework in terms of various types of information which can be used to build prediction models. Hence, it is important to develop techniques to include these datasets but in an \textit{End-to-End} fashion in order to minimize human effort and maximize performance.

\subsubsection{A Baseline method for incorporating road networks} \label{sec:baseline}

We present a baseline naive method which can be used to incorporate road networks to the \textit{Rule based End-to-End} framework without manual feature engineering. The method includes rasterizing of the road network with the same resolution as of the raster $R$ and then adding the newly formed raster as a new band to $R$. Subsequent to this, the same procedure is followed as in \textit{Rule based End-to-End} framework to get the urban growth model. This two-step addition to the framework is a trivial way of plugging road networks but this method drastically reduces the performance of \textit{End-to-End} framework (see Section \ref{sec:resultsdiscussion}). This is possibly because during rasterizing with a particular resolution, information is lost. Hence in this paper, we present a method with which we can improve over the previous framework.

\subsection{Recurrent Neural Networks} \label{sec:rnn}

Recurrent Neural Networks are a type of artificial neural networks where we have feedback from the output of a node back to the network. Unlike feed forward neural networks,  recurrent neural networks can remember information from previous executions thus giving an impression of memory (\cite{mitchell1997machine}). Hence these networks have the ability to process variable size sequence data and generate fixed length representations.

\begin{figure}
\centering
\includegraphics[scale=0.25]{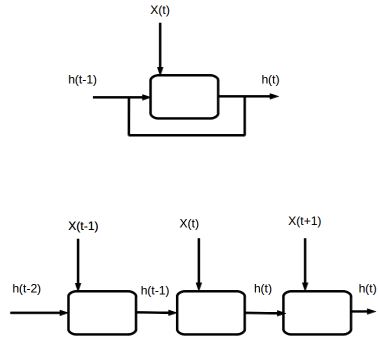}
\caption{Recurrent Neural Network}
\label{fig:rnn}
\end{figure}

Figure \ref{fig:rnn} depicts a schematic of a recurrent neural network. The variable $h_t$ in the diagram represents the memory of the recurrent neural network. A recurrent neural network can be unrolled with a fixed time steps as shown in the figure to form a feed forward network. Unrolling the network is useful during training the network as these networks are trained using backpropagation algorithm. Since multiple time steps are included in training the network, the algorithm often suffers from vanishing gradient problem (\cite{pascanu2013difficulty}).

The generalized equation of a recurrent neural network can be written as,

\begin{eqnarray}
h_t = f([h_{t-1}, X_t])
\label{eqn:rnn}
\end{eqnarray}

where  $h_t$ is the output as well as memory variable, $X_t$ is the input to the network and $f$ the function depicting a neural unit.

\subsection{Long Short Term Memory} \label{sec:lstm}

Long Short Term Memory (LSTM) is a specialized recurrent neural unit developed by \cite{hochreiter1997long} that deals with the vanishing gradient problem of recurrent neural networks. It consists of recurrent gates which can remember and drop information according to its parameters. The parameters of the \textit{remember} and \textit{forget} gates can be learned during the gradient descent algorithm. Due to this property, LSTM can learn tasks that require memory of events occurred in distant past (\cite{sutskever2014sequence}). 

\begin{figure}
\centering
\includegraphics[scale=0.25]{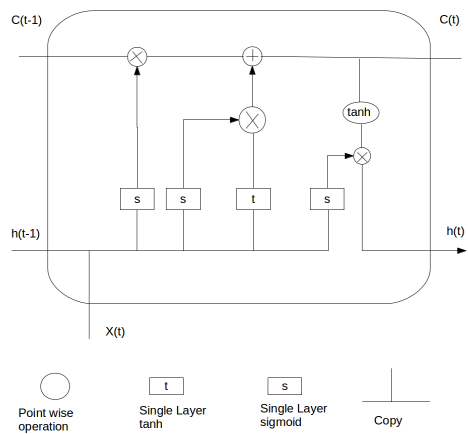}
\caption{A basic LSTM cell}
\label{fig:basiclstmcell}
\end{figure}

Figure \ref{fig:basiclstmcell} shows structure of a basic lstm cell. The LSTM cells consist of three sigmoid layers, a tanh layer, and four pointwise operations. The governing equations of the LSTM cell are as follows.

\begin{eqnarray}
C_t = (C_{t-1}\times \sigma([X_{t-1}, h_{t-1}])) + (\sigma([X_{t-1}, h_{t-1}])\times tanh([X_{t-1}, h_{t-1}]))
\label{eqn:longterm}
\end{eqnarray}

\begin{eqnarray}
h_t = \sigma([X_{t-1}, h_{t-1}])\times tanh(C_t)
\label{eqn:output}
\end{eqnarray}

Variables $C_t$ and $h_t$ in eqn. (\ref{eqn:longterm}) and (\ref{eqn:output}) are considered to be the long term and short term memory respectively. Equation (\ref{eqn:longterm}) determine which information to add/remove from $C_t$. From the information of the long term $C_{t-1}$, short term $h_{t-1}$ and current input $X_t$, we determine the output $h_t$. The paramters of the cell are adjusted with the input sequence data using stochastic gradient descent algorithm.

\subsection{Autoencoders} \label{sec:autoencoder}

Autoencoders are neural network architectures which consist of an encoder and a decoder and is used to generate representations in an unsupervised manner from data (\cite{vincent2010stacked}). Autoencoders encode an input vector to an encoded representation followed by decoding the representation back to a vector of the same size to that of the input vector. The parameters of the encoder and the decoder are estimated by minimizing the error between the decoded and the original vector. Hence, they are useful in pretraining a deep neural network with many layers when labeled data is not present or is not in sufficient quantity. 

The encoder and decoder can be mathematically represented as follows.

\begin{eqnarray}
p = f(W_e x + b)
\label{eqn:encoder}
\end{eqnarray}

\begin{eqnarray}
x' = f(W_d p + b_d)
\label{eqn:decoder}
\end{eqnarray}

The objective function optimization to find parameters $<W_e, W_d,b,b_d>$ are given as follows.

\begin{eqnarray}
min_\theta \frac{1}{N}\sum{|x'-x|^2}
\label{eqn:autoencoder_optmization}
\end{eqnarray}

In this paper, we have used the above components and concepts to develop an architecture to generate fixed length representations from the road network data and integrate into the \textit{End-to-End} framework. The addition of road network data to the framework improved the performance of the \textit{End-to-End} framework on urban growth prediction on the city of Mumbai.

\section{Methodology}

\begin{figure}
\centering
\includegraphics[scale=0.3]{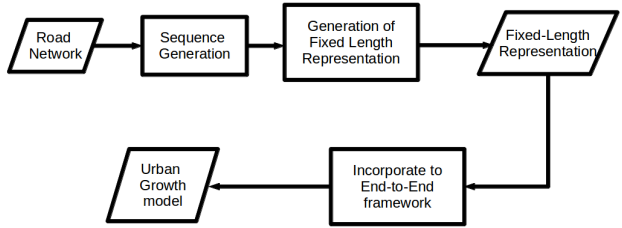}
\caption{Flowchart of the process of generation of fixed length representation}
\label{fig:fixed_length_rep}
\end{figure}

In this section, we discuss the methodology of generating fixed length representations from road network using a dynamic recurrent autoencoder and incorporating it into the \textit{End-to-End} framework. A flowchart of the procedure is shown in Fig. \ref{fig:fixed_length_rep}.

\begin{itemize}

\item The process uses road network data and generates a list of variable length sequences which we call \textit{Position Based Representation (PBR)} vector. The sequence is position based because it depends on the position of a cell in the cellular automata model as in eqn. \ref{eqn:ca_model_update} (\textit{Rule Based End-to-End framework}). 

\item From the generated variable-sized sequences, we generate fixed length representations which can be incorporated in the \textit{End-to-End} framework. Fixed length representations are important because most classifiers, like KNN, SVM, ANN, take fixed length feature vectors.  

\item Finally, we incorporate the fixed length representation of the sequences (\textit{PBR}) into the \textit{End-to-End} framework and build models.

\end{itemize}

\subsection{Sequence Generation from road networks} \label{sec:sequence_gen}

The sequence generation step of the process flowchart generates sequences from the road network such that it conforms to the cellular automata model in eqn. \ref{eqn:ca_model_update}. The updated eqn. \ref{eqn:ca_model_update} after adding a road network term is stated as

\begin{eqnarray}
\tau_p^{t+1} = f_T(l_p^t, N(l_p^t), \phi_{raster}^{len}(R_p, N(R_p)), \phi_{road}^{len}((reg(p)\cup reg(N(p)))\cap R_n))
\label{eqn:ca_model_update_roads}
\end{eqnarray}

, where $R_n$ denotes road network, $reg(p)$ denotes a geographical region represented by a cell or cells and $\phi_{road}^{len}$ is a function which returns a fixed length representation of length $len$ from a variable sized sequence. $\phi_{road}^{len}((reg(p)\cup reg(N(p)))\cap R_n)$ is a fixed length representation obtained from the road network. The term $(reg(p)\cup reg(N(p))$ is required because in the \textit{End-to-End} framework, we have assumed a neighborhood criterion $N()$ when assuming cellular automata behaviour which means a particular cell is only dependent on neighborhood cells. Inorder to fulfill this assumption, we generate a extract road fragments present in the neighborhood of a cell $p$ to form a \textit{Position Based Representation (PBR) Vector} (Discussed in Section \ref{sec:pbr}). 

An issue is the variable size of the roads in a road network due to which it is non-trivial to structure the data in a relational form. Hence, we have used a Dynamic Recurrent Neural Network Autoencoder (discussed in Section \ref{sec:drae}) architecture to transform road sequences to a fixed length representation. This is useful because fixed length representations can be effortlessly used as feature vectors of a model.

\subsubsection{Position Based Representation vector (PBR)} \label{sec:pbr}

\begin{figure}
\centering
\includegraphics[scale=0.25]{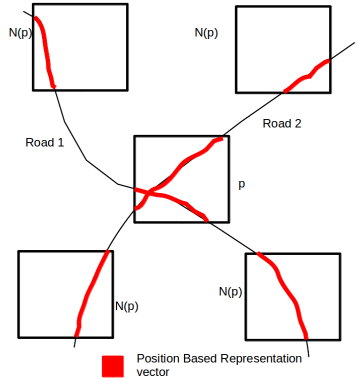}
\caption{Cell and its Neighborhood superimposed on a road network}
\label{fig:pbr}
\end{figure}

PBR vector is a portion of a road network which intersects with the region represented by a cell and its neighborhood in the cellular automata model of the \textit{End-to-End} framework. The definition indicates that \textit{PBR} vectors depend on a cell and the neighborhood criterion of the \textit{End-to-End} framework. A possible superimposition of cells of a cellular automata is depicted in Figure \ref{fig:pbr}. The cell size depends on how the cellular automata is defined but currently for brevity, we can assume them as pixels of the satellite image raster. The generation of \textit{PBR} vectors is given in Algorithm \ref{algo:sequence_gen}. 

Each road which intersects with a cell or its neighborhood will generate a \textit{PBR} vector. Hence there can be multiple \textit{PBR} vectors for a single point. For instance in Fig. \ref{fig:pbr}, in the rectangular box p intersects with both road 1 and 2. Hence, there will be two \textit{PBR} vectors.

The procedure to generate a list of \textit{PBR} vectors is given in Algorithm \ref{algo:sequence_gen} and is composed of two steps (two outer loops).

\begin{itemize}
\item The first part of the procedure uses two custom functions namely $coord2pixel$ and $pixel2bbox$, which can be easily developed if someone is using ESRI shapefile\footnote{http://www.esri.com/library/whitepapers/pdfs/shapefile.pdf} formats and standard Geotiff\footnote{https://trac.osgeo.org/geotiff/} raster formats. We consider a matrix of lists $H$ of size equal to that of the \textit{built-up} raster where we store vector ids where a road intersects with a pixel. $H$ works similar to a chained hash table\footnote{http://www.geeksforgeeks.org/hashing-set-2-separate-chaining/} for the next part in which we generate the \textit{PBR} vectors according to the neighborhood criterion. The hash table $H$ reduces the search time by storing the road vector ids $v$. During preparation of $H$ it is possible that two coordinates $c_i$ and $c_{i+1}$ fall on two distant points $p_i$ and $p_{i+1}$. Since the network is connected, therefore all the points falling in the straight line between $p_i$ and $p_{i+1}$ must also be connected. In order to achieve this we have used Bresenham's line drawing algorithm\footnote{http://www.idav.ucdavis.edu/education/GraphicsNotes/Bresenhams-Algorithm.pdf} ($line(p,q)$) to detect the points which fall between $p_i$ and $p_{i+1}$. 

\item The second part of the algorithm involves generation of bounding boxes $bbox$ from each pixel $P$ and its neighborhood $N(p)$ followed by performing spatial intersection $\cap_s$ with the road vectors of vector ids present in positions $\{p,N(p)\}$ of $H$. Spatial intersection ($\cap_s$) is an intersection of two geometries to produce another geometry, for instance, the spatial intersection of two non-parallel lines gives a point. 

\end{itemize}

\begin{algorithm}
\begin{algorithmic}
\STATE {\bf Input}
\STATE $B_t \gets$ Built-up raster at time $t$
\STATE $Rv \gets$ Road network Vector Data
\STATE $d \gets$ Number of decimal places to consider in a coordinate system
\STATE {\bf Variables}
\STATE $H \gets$ A two dimensional array of lists of size same as $B_t$
\STATE {\bf Custom functions}
\STATE $coord2pixel(c) \gets$ Function which inputs a coordinate and outputs a pixel.
\STATE $pixel2bbox(p, B_t) \gets$ Function to convert a pixel to its corresponding bounding box.
\STATE $line(p,q) \gets$ Function which uses Bresenham's line algorithm to find pixels falling on a line between $p$ and $q$
\STATE $\cap_s \gets$ Intersection of two geometries to provide another geometry (spatial intersection)
\STATE {\bf Output}
\STATE $Rx \gets$ List containing \textit{Position Based Representation (PBR) Vector} of Road network
\STATE {\bf Procedure}
\FORALL {vector $v \in Rv$}
	\FORALL {$\{c_i, c_{i+1}\}$ in $v$ }
		\STATE $p_i \gets coord2pixel(c_i)$
		\STATE $p_{i+1} \gets coord2pixel(c_{i+1})$
		\FORALL {$p \in line(p_i, p_{i+1})$ }
			\IF {$v \notin H[p]$}
				\STATE $H[p] \gets H[p] \cup v_{id}$
			\ENDIF
		\ENDFOR
	\ENDFOR
\ENDFOR
\FORALL {$p\in H$}
	\STATE $bbox \gets pixel2bbox(q), \forall q \in \{p,N(p)\}$
	\FORALL {$v \in H[p]$}
		\STATE $temp_{list} \gets v\cap_s bbox$ // $\cap_s$ is spatial intersection
		\STATE $temp_{list} \gets \{x_p, y_p\}, \forall p \in temp_{list}$ 
		\STATE $Rx \gets Rx\cup temp_{list}$
	\ENDFOR
\ENDFOR
\STATE return $Rx$
\end{algorithmic}
\caption{Sequence Generation Procedure}
\label{algo:sequence_gen}
\end{algorithm}

\begin{algorithm}
\begin{algorithmic}
\STATE {\bf Input}
\STATE $B_t \gets$ Built-up raster at time $t$
\STATE $Rx \gets$ List containing \textit{Position Based Representation (PBR) Vector} of Road network
\STATE $epochs \gets$ Number of Epochs of training
\STATE $rep_{size} \gets$ Size of the fixed length representation
\STATE {\bf Variables}
\STATE $drnnae \gets \{lstm_{enc}, lstm_{dec}, rep_{size}\}$ Dynamic Recurrent Neural Network Autoencoder
\STATE $sdrnnae \gets \{drnnae_{lat}, drnnae_{lon}\}$ Spatial Dynamic Recurrent Neural Network Autoencoder
\STATE {\bf Custom functions}
\STATE $mean(matrix, axis) \gets$ Function to compute mean of elements on a particular axis (row-wise or column-wise)
\STATE $train(autoencoder, fv) \gets $ Function to train an autoencoder with feature vectors $fv$
\STATE {\bf Output}
\STATE $R_{flv}^p \gets$Fixed length Representation vector for pixel $p$ in $B_t$
\STATE {\bf Procedure}
\FOR {$i\gets 1 \textit{ to } epochs$}
	\FORALL {$batch_x \in Rx$}
		\STATE $train(sdrnnae, batch_x)$
	\ENDFOR
\ENDFOR
\STATE $R_{flv}\gets lstm_{enc}.encode(Rx)$
\FORALL {$p \in B_t$}
	\STATE $R_{flv}^p \gets mean(r, axis=0), \forall r \in R_{flv}^p $
\ENDFOR
\STATE return $R_{flv}$
\end{algorithmic}
\caption{Fixed length Representation generation}
\label{algo:fixed_length_representation}
\end{algorithm}

\subsection{Dynamic Recurrent Neural Network Autoencoder (DRNNAE)} \label{sec:drae}

\begin{figure}
\centering
\includegraphics[scale=0.3]{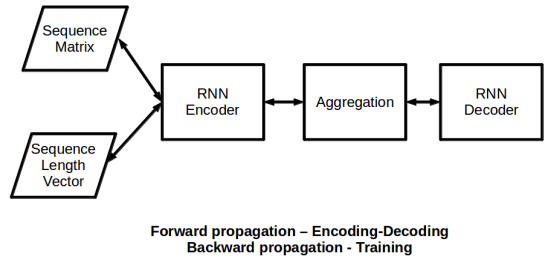}
\caption{Dynamic Recurrent Neural Network Autoencoder}
\label{fig:drnnae}
\end{figure}

Dynamic Recurrent Neural Network Autoencoder is a recurrent neural network which can learn representations from variable length sequences in an unsupervised manner. Figure \ref{fig:drnnae} shows the diagram of the autoencoder. The architecture is composed of two components namely, an RNN encoder and decoder. The encoder encodes the features, whereas the decoder decodes it back to the original input which is similiar to an Autoencoder (discussed in Section \ref{sec:autoencoder}). This assists us in training the architecture in an unsupervised manner. 

From Section \ref{sec:rnn}, we have seen that a recurrent neural network can be unrolled into a number of basic cells. Due to this property, a recurrent neural network can unroll itself into variable length according to a sequence. For instance, for a sequence of length twenty, we unroll the RNN for twenty steps and for a sequence of length thirty in the same dataset, we unroll the RNN thirty times. In each of the cases, the output $h_t$ from eqn. (\ref{eqn:rnn}), is of fixed dimensionality. Thus, the recurrent neural network is capable of generating fixed sized representations from variable sized sequences in the same dataset.

In practice, recurrent neural networks face a problem during training which is called the vanishing gradient problems \cite{sutskever2013training}. In order to avert the problem, we have used a specialized recurrent basic cell which is the LSTM (discussed in Section \ref{sec:lstm}). LSTM is capable of running on sequences of long sizes because of the unique ability to remember long-term as well as short-term information (\cite{lecun2015deep}). This makes it suitable for learning long as well as short sized variable length sequences.

\begin{figure}
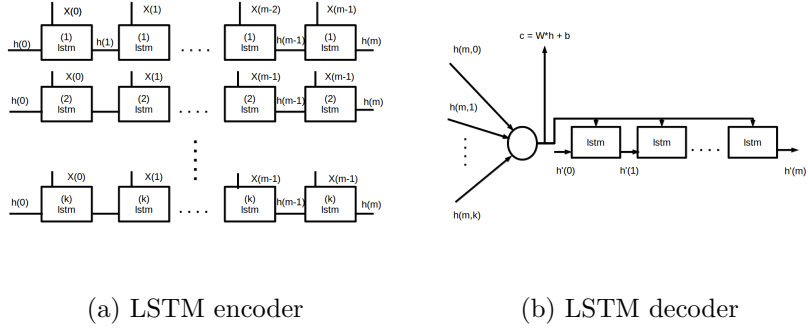

\centering
\subfloat[][LSTM encoder]{\includegraphics[scale=0.2]{\locimage{lstm_encoder}}}
\subfloat[][LSTM decoder]{\includegraphics[scale=0.2]{\locimage{lstm_decoder}}}
\caption{Dynamic Recurrent Neural Network (a) Encoder (b) Decoder}
\label{fig:drnnencdec}
\end{figure}

The architecture is composed of an encoder and a decoder, descriptions of which are given as follows.

\begin{itemize}

\item Figure \ref{fig:drnnencdec}(a) illustrates the RNN Encoder architecture. There are $k$ streams of LSTM sequences which ultimately results in a $k$ sized representation of the input vector $X$ of size $m$ (provided that the LSTMs have different parameters). It may be noted that $m$ can vary according to the size of $|X|$. Hence the guiding equation for the encoder is given as follows.

\begin{eqnarray}
C_t^k, h_t^k = f^k_{lstm}(h_{t-1}, X_t, C_{t-1})
\label{eqn:encode}
\end{eqnarray}

\item Figure \ref{fig:drnnencdec}(b) illustrates the RNN Decoder architecture. The $k$ sized representation by the LSTM sequences are used to generate $m$ sized sequences. The guiding equation for the decoder is as follows.

\begin{eqnarray}
C_t', h_t' = f_{lstm}(h_{t-1}', Wh_t+b, C_{t-1}')
\label{eqn:decode}
\end{eqnarray}

\item The objective function for estimation of parameters is given as follows.

\begin{eqnarray}
E = \frac{1}{N} \sum{|h_t' - X_t|^2}
\label{eqn:optimize}
\end{eqnarray}

\end{itemize}

The procedure for generation of fixed length representation is given in Algorithm \ref{algo:fixed_length_representation}. The algorithm trains a SDRNNAE (discussed in Section \ref{sec:sdrnnae}) which is composed of two DRNNAE because the \textit{PBR} vectors have two dimensions viz latitude and longitude. It involves training the autoencoder using stochastic gradient descent and then generating encoded feature vectors. Since for certain cases, we can have more than one fixed length vectors, we resolve it by taking mean of the vectors row-wise.

\begin{algorithm}
\begin{algorithmic}
\STATE {\bf Non-trainable Parameters}
\STATE $lseq_{max} \gets $ Maximum length of sequence 
\STATE $n_{hidden}\gets $ Number of hidden LSTM units
\STATE $S_m \gets $ Length of $m^{th}$ sequence 
\STATE {\bf Trainable units and parameters}
\STATE $f_{lstm} \gets$ A Basic cell of LSTM as shown in Fig. \ref{sec:lstm}
\STATE $[W, B] \gets$ Weights and Biases.
\STATE {\bf LSTM encoder for latitude}
\STATE $C_t^k, h_t^k \gets f^{lat}_{lstm}(h_{t-1}^k, X_m^t, C_{t-1}^k)$, $\forall k \in {1,2,3,.. n_{hidden}}$ and $t \in \{1,2,..S_m\}$
\STATE {\bf LSTM decoder for latitude}
\STATE $c \gets W_e^{lat}h + B_e^{lat}$
\STATE $C_t', h_t' \gets f^{lat}_{lstm}(h_{t-1}', c, C_{t-1}')$ and $t \in \{1,2,3,...S_m\}$
\STATE $X_m' \gets W_d^{lat}h'$
\STATE {\bf Minimization criterion}
\STATE $E \gets \frac{1}{S_m} \sum{|X_m' - X_m|^2}$
\STATE {\bf LSTM encoder for longitude}
\STATE $C_t^k, h_t^k \gets f^{lon}_{lstm}(h_{t-1}^k, Y_m^t, C_{t-1}^k)$, $\forall k \in {1,2,3,.. n_{hidden}}$ and $t \in {1,2,..S_m}$
\STATE {\bf LSTM decoder for longitude}
\STATE $c \gets W_e^{lon}h + B_e^{lon}$
\STATE $C_t', h_t' \gets f_{lstm}^{lon}(h_{t-1}', c, C_{t-1}')$ and $t \in \{1,2,3,...S_m\}$
\STATE $Y_m' \gets W_d^{lon}h'$
\STATE {\bf Minimization criterion}
\STATE $E \gets \frac{1}{S_m} \sum{|Y_m' - Y_m|^2}$
\end{algorithmic}
\caption{Spatial Dynamic Recurrent Neural Network Autoencoder}
\label{algo:sdrnnae}
\end{algorithm}

\subsubsection{Spatial Dynamic Recurrent Neural Network Autoencoder} \label{sec:sdrnnae}

In this work, we have used two \textit{DRNNAE} for representation two axes of the coordinate system to generate fixed length representation. The description of the \textit{Spatial Dynamic Recurrent Neural Network Autoencoder} in Algorithm \ref{algo:sdrnnae}. There are two encoders and decoders which are trained separately for sequences of $latitude$ and $longitude$ coordinates. The model generates fixed length representations of size $2n_{hidden}$. The algorithms of training recurrent neural networks can be obtained from several other works (\cite{sutskever2013training, haykin2009neural}).

\subsection{Features of the Dynamic Recurrent Neural Network Autoencoder}

The proposed architecture for generating fixed length representation from road networks has the following qualities.

\begin{itemize}

\item The algorithms for generation of fixed length representations are unsupervised. There is no requirement of any manual labeling in application of both the algorithms (Algorithm \ref{algo:sequence_gen} and \ref{algo:fixed_length_representation}). This is useful as there are large repositories of transportation data available which can be utilized for prediction purposes without manual labeling or feature engineering. 

\item The algorithms generate representations which are of fixed size. Fixed sized representations have enormous utility because they can be used in machine learning models as training data to find patterns. Hence given a representation, different applications can be build using it, thus making the modeling process robust.

\item No manual feature engineering is involved in the process of generation of fixed length representation. Hence, the process is generic and depends largely on the data provided and the hyperparameters of the training process.

\end{itemize}

Finally, the architecture comes with a drawback. Since the representations generated by the architecture are numerical in nature, therefore it provides us no clue as to what it means. Hence, evaluation is only limited to visualization and various kinds of accuracy metrics for different applications. Furthermore, since the network can go to large depths, computation time for estimating parameters of the encoder as well as encoding process increases.

\section{Experiments and Results}

In this section, we provide details of the experiments and inferences which have been carried out on the region of Mumbai (latitude: $19.0760\degree$ N, longitude: $72.8777\degree$ E). Mumbai is the capital city of Maharashtra and one of the major cities of India which has seen urbanization in last $20$ years according to the Census of India. According the reports by the Census, population of Mumbai has steadily risen from approximately $9$ million in $1991$ to more than $12$ million in $2011$. 

The experiments have been conducted in a \textit{Virtual Machine (VM)} running in an \textit{Open Stack} based cloud infrastructure. The VM consists of \textit{8 VCPUs}, \textit{16 GB RAM} and \textit{Ubuntu 14.04} as operating system.

\subsection{Data collection and Preprocessing}

\begin{figure}
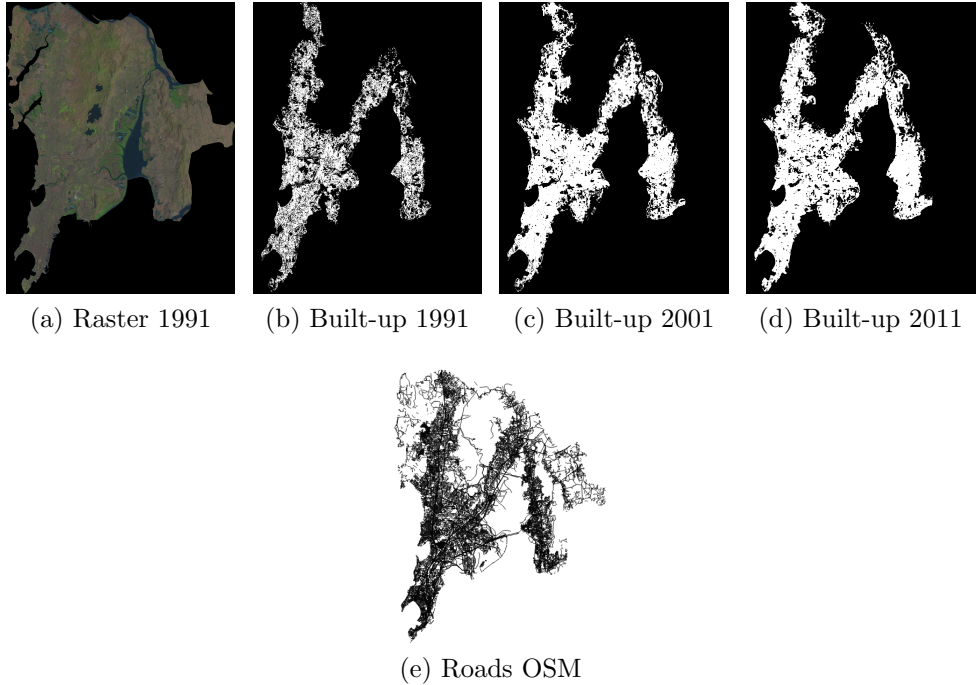

\centering
\subfloat[][Raster 1991]{\includegraphics[scale=0.07]{\locimage{mumbai1991}}}\hspace{0.1cm}
\subfloat[][Built-up 1991]{\includegraphics[scale=0.07]{\locimage{cimg1991}}}\hspace{0.1cm}
\subfloat[][Built-up 2001]{\includegraphics[scale=0.07]{\locimage{cimg2001}}}\hspace{0.1cm}
\subfloat[][Built-up 2011]{\includegraphics[scale=0.07]{\locimage{cimg2011}}}\hspace{0.1cm}
\subfloat[][Roads OSM]{\includegraphics[scale=0.2]{\locimage{roadsmumbai}}}
\caption{Dataset of the Mumbai, India area}
\label{fig:dataset_mumbai}
\end{figure}

We have two kinds of data requirement for performing experiments -- one is raster images for year $1991$, $2001$ and $2011$ and road network data. The raster images are downloaded from United States Geological Survey (USGS) website and the road network data for the region of Mumbai have been collection from the Open street maps data portal\footnote{https://www.openstreetmap.org/}. The extracted region of Mumbai from the satellite image consists of $972280$ pixels. It is important to ensure whether both the data are following the same coordinate reference frame before starting experiments. From the raster images of three time steps we have derived the urban built-up maps. The segmentation method is maximum likelihood classification which is a semi-automatic method of classification of the raster images. The generated maps are verified using Google timelapse\footnote{https://earthengine.google.com/timelapse/}, Google Maps\footnote{https://www.google.com/maps} as well as previously published works on Mumbai \cite{moghadam2013spatiotemporal, shafizadeh2015performance, rienow2015supporting}. The number and percentages of transition and persistent pixels are shown in Table \ref{table:pixel_transformed}. The raster, built-up images and road network are shown in Fig. \ref{fig:dataset_mumbai}.

\begin{table}
\centering
\begin{tabular}{|c|c|c|}
\hline
Time step & $\%$ transformed & $\%$ persistent\\
\hline
$1991-2001$ & $15\%$ & $85\%$\\
\hline
$2001-2011$ & $10\%$ & $90\%$\\
\hline
\end{tabular}
\caption{Number of pixels transformed vs number of pixels persistent}
\label{table:pixel_transformed}
\end{table}

\begin{figure}
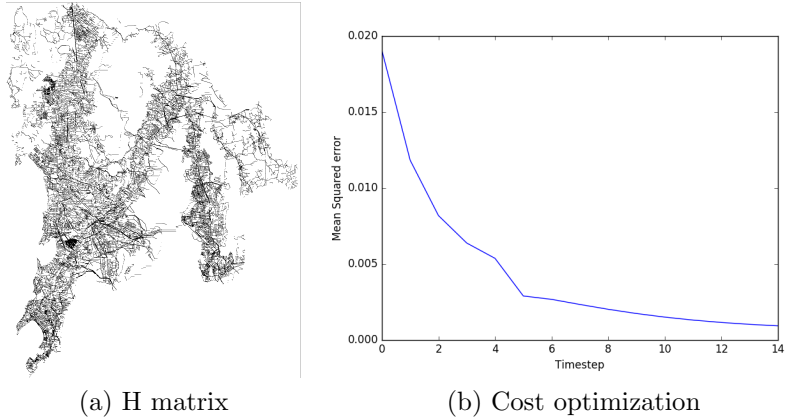

\centering
\subfloat[][H matrix]{\includegraphics[scale=0.09]{\locimage{structure_H}}}\hspace{0.1cm}
\subfloat[][Cost optimization]{\includegraphics[scale=0.33]{\locimage{cost_optimization}}}
\caption{(a) Visualization of the $H$ matrix. White pixels indicate nonzero list of road vectors. (b) Cost optimization of the SDRNNAE.}
\label{fig:visualize_H}
\end{figure}

The preprocessing consists of the generation of \textit{PBR} vectors from the road network data by employing Algorithm \ref{algo:sequence_gen}. As discussed in Section \ref{sec:sequence_gen}, the algorithm comprises formation of a matrix $H$ of lists of vectors and thereby generating \textit{PBR} from the road network. The structure $H$ is a crucial data structure which has the same number of rows and columns as the image raster and each position is a list of vectors of roads which passes through that position. In figure \ref{fig:visualize_H}(a), we depict a visualization  of the matrix $H$ by generating a binary image where a non-white pixel at a position indicates a nonempty list of vectors. The structure $H$ is important as it reduces the time for generating the \textit{PBR} vectors by acting as a lookup table.

From the structure $H$, Algorithm \ref{algo:sequence_gen} generates \textit{PBR} vectors. We have considered neighborhood criterion $N()$ in Algorithm \ref{algo:sequence_gen} as Moore neighborhood of radius $1$ which gives the maximum length of a \textit{PBR} vector as $18$. This means, that the Dynamic Recurrent Neural Network Autoencoder has to unroll at maximum up to $36$ time steps. If we take in terms of a feed forward neural network, then this equivalent to a $36$-layer feed forward neural network. Hence, we can also call this architecture a deep architecture.

\subsection{Training and Representation}

The list of sequences generated from the previous step has been used for training a Dynamic Recurrent Neural Network Autoencoder (Section \ref{sec:drae}). The number of sequences generated from the road network is equal to $1181974$. We have used stochastic gradient descent with batches of size $100000$ to train the autoencoder by varying the number of time steps and size of fixed length representation with a learning rate of $0.1$. After the training process is done, the trained architecture is used to generate a fixed length representation (see Algorithm \ref{algo:fixed_length_representation}). The architecture has been developed using \textit{tensorflow}\footnote{https://www.tensorflow.org/} library of python. 

For validation of unsupervised methods, there are two kinds of approaches namely, intrinsic and extrinsic. Extrinsic methods use available ground truth whereas intrinsic methods do not use any ground truth data for validation (\cite{han2011data}). For extrinsic validation, we have used the validation criteria of urban growth ($FoM, PA, UA$ and $OA$) to quantitatively measure the representation quality. For intrinsic verification, we have performed qualitative (visual) verification of reconstructed \textit{PBR} from the original \textit{PBR}. 

The four validation metrics ($FoM, PA, UA$ and $OA$) by (\cite{pontius2008comparing}) are dependent on 5 variables given as follows.

\begin{itemize}
\item $A$ = Area of error due to observed change predicted as persistence.

\item $B$ = Area correct due to observed change predicted as change.

\item $C$ = Area of error due to observed change predicted in the wrong gaining category.

\item $D$ = Area of error due to observed persistence predicted as change.

\item $E$ = Area correct due to observed persistence predicted as persistence.
\end{itemize}

Figure of Merit ($FoM$) provides us the amount of overlap between the observed and predicted change. Producer's accuracy ($PA$) gives the proportion of pixels that model accurately predicted as change, given that the reference maps indicate observed change. User's accuracy ($UA$) gives the proportion of pixels that the model predicts accurately as change, given that the model predicts change. The equations of the metrics are given as follows.

\begin{eqnarray}
FoM = \frac{B}{A+B+C+D}
\label{eqn:FoM}
\end{eqnarray}

\begin{eqnarray}
PA = \frac{B}{A+B+C}
\label{eqn:PA}
\end{eqnarray}

\begin{eqnarray}
UA = \frac{B}{B+C+D}
\label{eqn:UA}
\end{eqnarray}

\begin{eqnarray}
OA = \frac{B+E}{A+B+C+D+E}
\label{eqn:OA}
\end{eqnarray}

\begin{figure}
\centering
\includegraphics[scale=0.6]{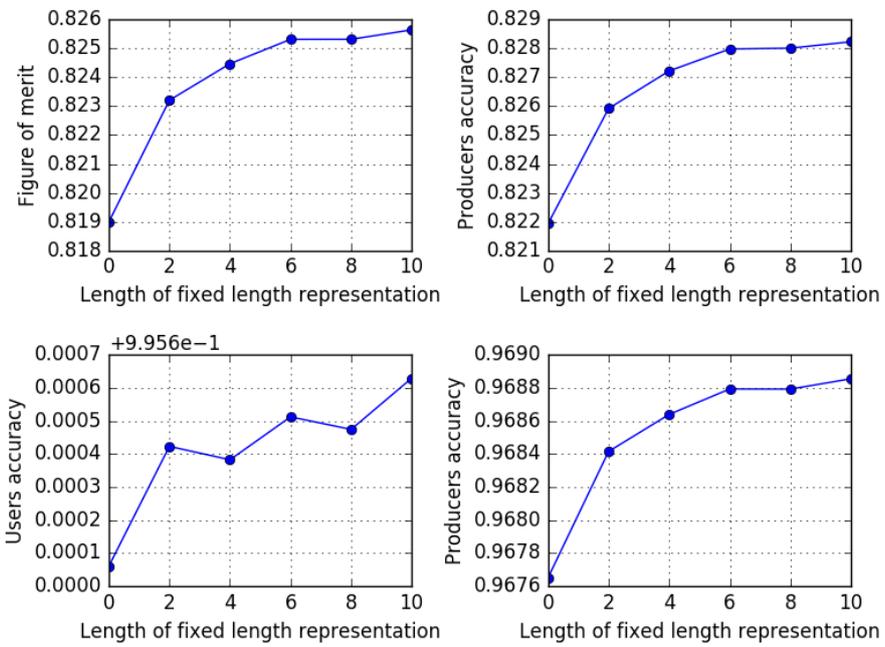}
\caption{Variation of metrices with length of fixed length representation -- Mumbai 1991-2001.}
\label{fig:sizefeature1991}
\end{figure}

\begin{figure}
\centering
\includegraphics[scale=0.6]{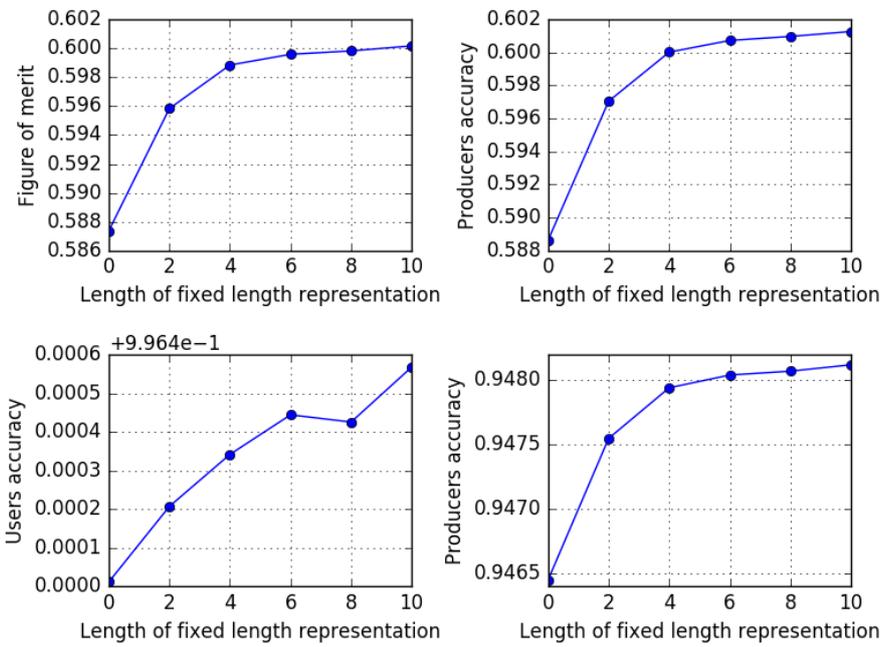}
\caption{Variation of metrices with length of fixed length representation -- Mumbai 2001-2011.}
\label{fig:sizefeature2001}
\end{figure}

We have validated our representation technique based on how these four metrics vary as the size of the fixed length representation and the number of iteration is varied. There are three divisions of experiments based on the size of feature count which is shown in Fig. \ref{fig:sizefeature1991} and \ref{fig:sizefeature2001}. We also determine the effectiveness of the training algorithm by checking the value of the objective function (eqn. \ref{eqn:optimize}) as the training proceeds. The cost optimization graph is given in Fig. \ref{fig:visualize_H}(b).

\subsection{Results and Discussion} \label{sec:resultsdiscussion}

The results presented in Fig. \ref{fig:sizefeature1991} and \ref{fig:sizefeature2001} indicates that there is a rise in the performance metrics (eqn. \ref{eqn:FoM}, \ref{eqn:PA}, \ref{eqn:UA}, and \ref{eqn:OA}) with increase in length of fixed length feature vectors. However, the rate of increase is not constant and reduces as the feature length is increased. This is possibly due to the increase in the number of dimensions in the data which brings in the issue of \textit{curse of dimensionality}. In high dimensional spaces, a finite dataset consisting of a fixed number of points turns sparse due to which predictive power of machine learning models reduces (\cite{hughes1968mean}). 

\begin{figure}
\centering
\subfloat[][1991-2001]{\includegraphics[scale=0.5]{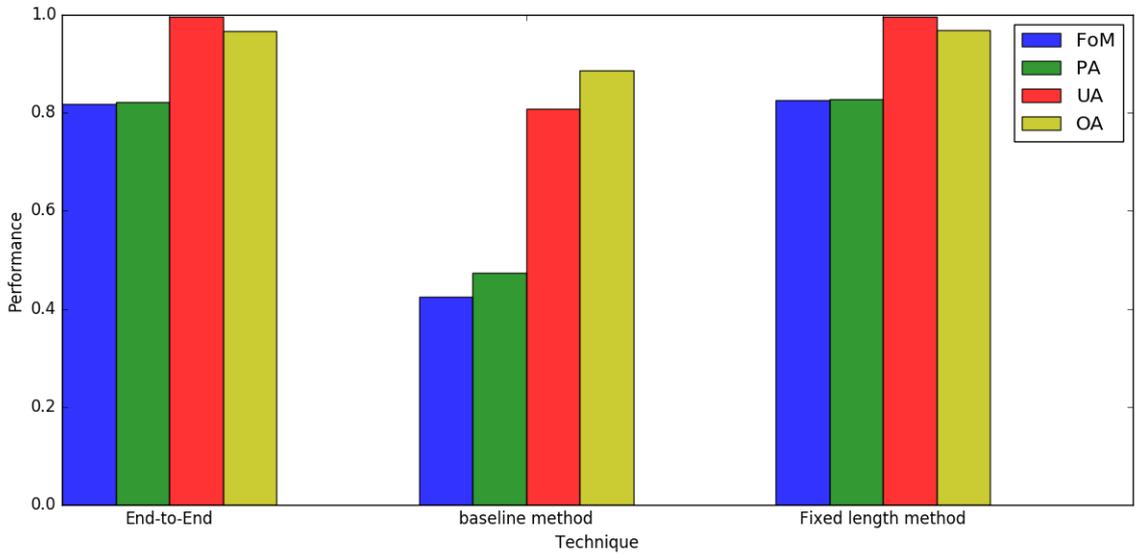}}

\subfloat[][2001-2011]{\includegraphics[scale=0.5]{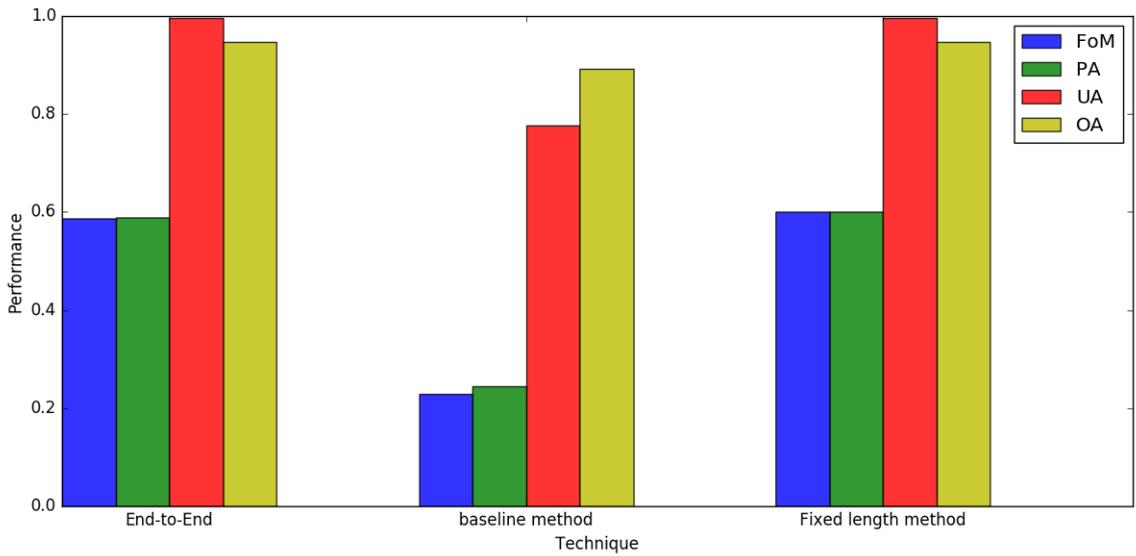}}
\caption{Performance analysis of fixed length representation with baseline and existing method.}
\label{fig:performance}
\end{figure}

On comparison of improvement of the fixed length representation technique with the \textit{baseline method}, as discussed in Section \ref{sec:baseline}, we find that our method performs better (Figure \ref{fig:performance}). In the baseline method, the performance metrics decreases than the former \textit{Rule based End-to-End} framework, whereas our method gives enhanced performance. This indicates that our method is generating meaningful representations from the road network data. A summary of improvements over baseline method is stated below.

\begin{itemize}

\item $20\%$ improvement in Figure of merit.

\item $17\%$ improvment in Producer's accuracy.

\item $10\%$ improvement in User's accuracy.

\item $3\%$ improvment in Overall accuracy.

\end{itemize}

A summary of improvement over the existing \textit{Rule based End-to-End} framework on performance metrics is given as follows.

\begin{itemize}

\item $0.8\%$ improvement in Figure of Merit.

\item $0.9\%$ improvment in Producer's accuracy.

\item $0.6\%$ improvement in User's accuracy.

\item $0.2\%$ improvment in Overall accuracy.

\end{itemize}

The baseline method works poorly due to the rasterization of the road network which causes loss of much of the spatial information present in the network. Contrary to that in our method, the spatial information is preserved in the encoding done by the SDRNNAE, the extent of which depends on the encoding size. Due to this, we see improvement in the performance metrics. However, the increase is not arbitrary and is limited by the resolution of the built-up image data which we have generated from the satellite imagery. Since road network vector data is a finer source of data, therefore to fully recognize its capability it is important to have a finer resolution of built-up information, which is not available to us currently.

\begin{figure}
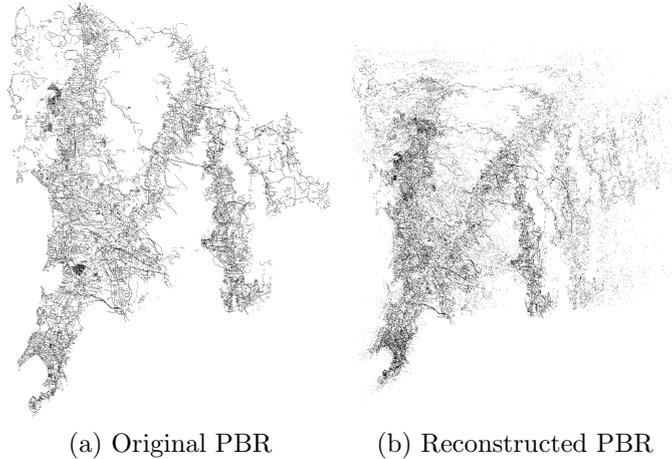

\centering
\subfloat[][Original PBR]{\includegraphics[scale=0.1]{\locimage{original_pbr}}}\hspace{0.1cm}
\subfloat[][Reconstructed PBR]{\includegraphics[scale=0.1]{\locimage{reconstructed_pbr}}}
\caption{Visualization of \textit{PBR} vectors -- original vs reconstructed after training.}
\label{fig:viz_pbr}
\end{figure}

Figure \ref{fig:viz_pbr} shows an extract from the original \textit{PBR} and the reconstructed PBR. The reconstruction has been carried out from encoded representations of size $10$. The reconstruction is crude but it gives us an idea that the LSTM decoder is able to decode the representations from the encoded representations by the LSTM encoder. We have used this to qualitatively verify our training process and hence can be considered as an intrinsic validation process. In short, reconstruction gives us an idea of the goodness of the representation.

\section{Conclusion}

In this paper, we proposed a method for generating fixed length representations to incorporate road network to \textit{Rule based End-to-End} framework. By incorporating road network using our method, we have observed a rise in the performance metrics of urban growth. However, the algorithms for generation of fixed length representation are time-consuming. Moreover, comprehending the fixed length representations is a challenging task. Further work in this area can be done on addressing these challenges as well as incorporating other forms of vector data as in point and polygon data.

\newcommand{\noopsort}[1]{} \newcommand{\printfirst}[2]{#1}
  \newcommand{\singleletter}[1]{#1} \newcommand{\switchargs}[2]{#2#1}

\end{document}